\documentclass{article}

\usepackage[square,numbers]{natbib}
\usepackage[utf8]{inputenc} 
\usepackage[T1]{fontenc}    
\usepackage{url}            
\usepackage{booktabs}       
\usepackage{amsfonts}       
\usepackage{nicefrac}       
\usepackage{microtype}      
\usepackage{lipsum}
\usepackage{tikz}
\usepackage{url,lineno,microtype,subcaption}
\usepackage{amsmath, amsxtra, amsfonts, amssymb, amstext, amsthm}
\usepackage{mathtools}
\usepackage{graphicx}
\usepackage[preprint]{neurips_2019}
\usepackage{wrapfig}

\newcommand{\fm}[1]{\comment{\textcolor{blue}{FM: #1}}}

\newcommand{\ls}{\langle}
\newcommand{\rs}{\rangle}
\newcommand{\tri}{\bigtriangledown}
\newcommand{\htri}{\hat{\bigtriangledown}}
\newcommand{\heps}{\hat{\epsilon}}
\newcommand{\hzeta}{\hat{\zeta}}
\newcommand{\eps}{\epsilon} 

\newtheorem{theorem}{Theorem}
\newtheorem{proposition}{Proposition}

\title{Improving Gradient Estimation in Evolutionary Strategies With Past Descent Directions}

\author{
	Florian Meier \thanks{Equal contribution.}\\
	Department of Computer Science\\
	ETH Zürich, Switzerland\\
	\texttt{meierflo@inf.ethz.ch} \\
	\And
	Asier Mujika \thanks{Author was supported by grant no. CRSII5\_173721  of the Swiss National Science Foundation.} \footnotemark[1] \\
	Department of Computer Science\\
	ETH Zürich, Switzerland\\
	\texttt{asierm@inf.ethz.ch} \\
	\And
	Marcelo Matheus Gauy \\
	Department of Computer Science\\
	ETH Zürich, Switzerland\\
	\texttt{marcelo.matheus@inf.ethz.ch} \\
	\And
	Angelika Steger \\
	Department of Computer Science\\
	ETH Zürich, Switzerland\\
	\texttt{steger@inf.ethz.ch} \\
}

\begin{document}
\maketitle

\begin{abstract}
Evolutionary Strategies (ES) are known to be an effective black-box optimization technique for deep neural networks when the true gradients cannot be computed, such as in  Reinforcement Learning. We continue a recent line of research that uses surrogate gradients  to improve the gradient estimation of ES.
We propose a novel method to optimally incorporate surrogate gradient information. 
Our approach, unlike previous work, needs no information about the quality of the surrogate gradients and is always guaranteed to find a descent direction that is better than the surrogate gradient.
This allows to iteratively use the  previous gradient estimate as surrogate gradient for the current search point. We theoretically prove that this yields fast convergence  to the true gradient for linear functions and show under simplifying assumptions that it significantly improves gradient estimates for general functions. Finally, we evaluate our approach empirically on MNIST and reinforcement learning tasks and show that it considerably improves  the gradient estimation of ES at no  extra computational cost.
\end{abstract}


\section{Introduction}
Evolutionary Strategies (ES) \cite{rechenberg1973evolution, schwefel1977evolutionsstrategien, nesterov2017random} are a black-box optimization technique, that estimate the gradient of some  objective function with respect to the parameters by evaluating parameter perturbations in random directions. The benefits of using ES in Reinforcement Learning (RL) were exhibited in \cite{salimans2017evolution}.
ES approaches are highly parallelizable and account for robust learning, while having decent data-efficiency. Moreover, black-box optimization  techniques like ES do not require propagation of gradients, are tolerant to long time horizons, and do not suffer from sparse reward distributions~\cite{salimans2017evolution}.  This lead to a successful application of ES in variety of different RL settings~\cite{choromanski2018structured,cui2018evolutionary,houthooft2018evolved, ha2018recurrent}. Applications of ES outside RL include for example  meta learning~\cite{metz2018learned}.

In many scenarios, the true gradient is impossible to compute, however surrogate gradients are available. Here, we use the term  \emph{surrogate gradients} for directions  that are correlated but usually not equal to the true gradient, e.g.\ they might be biased or unbiased approximations of the gradient. Such scenarios include models with discrete stochastic variables~\cite{bengio2013estimating}, learned models in RL like Q-learning~\cite{watkins1992q}, truncated backpropagation through time~\cite{rumelhart1985learning} and feedback alignment~\cite{lillicrap2014random}, see~\cite{maheswaranathan2019guided} for a detailed exhibition.
If surrogate gradients are available, it is beneficial to preferentially sample parameter perturbations from the subspace defined by these directions~\cite{maheswaranathan2019guided}.
The proposed algorithm~\cite{maheswaranathan2019guided} requires knowing in advance the quality of the surrogate gradient, does not always provide a descent direction that is better than the surrogate gradient, and it remains open how to obtain such surrogate gradients in general settings.

In deep learning in general, experimental evidence has established that higher order derivatives are usually "well behaved", in which case gradients of consecutive parameter updates correlate and applying momentum speeds up convergence~\cite{dozat2016incorporating, sutskever2013importance, ruder2016overview}. These observations suggest that past update directions are promising candidates for surrogate gradients.

In this work, we extend the line of research of~\cite{maheswaranathan2019guided}. Our contribution is threefold:
\begin{itemize}
    \item First, we show theoretically how to optimally combine the surrogate gradient directions with  random search directions. More precisely, our approach computes the direction of the subspace spanned by the evaluated search directions that is most aligned with the true gradient.
    Our gradient estimator does not need to know the quality of the surrogate gradients and always provides a descent direction that is more aligned with the true gradient than the surrogate gradient.
    \item Second, above properties of our gradient estimator allow us to iteratively use the last update direction as a surrogate gradient for our gradient estimator. Repeatedly using the last update direction as a surrogate gradient will aggregate information about the gradient over time and results in improved  gradient estimates. In order to demonstrate how the gradient estimate improves over time, we prove  fast convergence to the true gradient for linear functions and show, that under simplifying assumptions, it offers an improvement over ES that depends on the Hessian for general functions.
    \item Third, we validate experimentally that these results transfer to practice, that is, the proposed approach computes more accurate gradients than standard ES. We observe that our algorithm considerably improveŝ gradient estimation on the MNIST task compared to standard ES and that it improves  convergence speed and performance on the tested Roboschool reinforcement learning environments. 
\end{itemize}

\section{Related Work}

Evolutionary strategies  \cite{rechenberg1973evolution, schwefel1977evolutionsstrategien, nesterov2017random} are black box optimization techniques that approximate the gradient  by sampling finite differences in random directions in parameter space. 
Promising potential of ES for the optimization of neural networks used for RL was demonstrated in \cite{salimans2017evolution}. They showed that ES gives rise to efficient training despite the noisy gradient estimates that are generated from a much smaller number of samples  than the dimensionality of parameter space. This placed ES on a prominent spot in the RL tool kit~\cite{choromanski2018structured,cui2018evolutionary,houthooft2018evolved, ha2018recurrent}. 

The history of descent directions was previously used to adapt the search distribution in covariance matrix adaptation ES (CMA-ES) \cite{hansen2016cma}. 
CMA-ES constructs a second-order model of the underlying objective function and samples search directions and adapts step size according to it. However, maintaining the full covariance matrix makes the algorithm quadratic in the number of parameters, and thus impractical for high-dimensional spaces. Linear time approximations of CMA-ES like diagonal approximations of the covariance matrix \cite{wierstra2014natural} often do not work  well. E.g.\ even for linear functions the gradient estimates will not converge to the true gradient, instead the step-size of the descent direction is arbitrarily increased.
Our approach differs  as we simply improve the gradient estimation and then feed the gradient estimate to a first-order optimization algorithm.

Our work is inspired by the line of research of~\cite{maheswaranathan2019guided}, where  surrogate gradient directions are used to improve gradient estimations by 'elongating' the search space along these directions. That approach has two  shortcomings. First, the bias of the surrogate gradients needs to be known to adapt the covariance matrix. Second, once the bias of the surrogate gradient is too small, the algorithm will not find a better descent direction than the surrogate gradient. 

Another related area of research investigates how to use momentum for the optimization of deep neural networks. Applying different kinds of momentum has become one of the standard tools in current deep learning and it has been shown to speed-up learning in a very wide range of tasks \cite{kingma2014adam, sutskever2013importance, ruder2016overview}. 
This hints, that for many problems the higher-order terms in deep learning models are "well-behaved" and thus, the gradients do not change too drastically  after parameter updates. While these approaches use  momentum for parameter updates,  our approach can be seen as a form of momentum when sampling directions from the search space of ES.

\section{Gradient Estimation}
We aim at minimizing a function $f: \mathbb{R}^n \rightarrow \mathbb{R}$ by steepest descent. In scenarios where the gradient $\tri f$ does not exist or is inefficient to compute, we are interested in obtaining some estimate of the (smoothed) gradient of $f$ that provides a good parameter update direction.

\subsection{The ES Gradient Estimator}
ES considers the function $f_{\sigma}$ that is obtained by \emph{Gaussian smoothing}
\begin{equation*}
    f_{\sigma}(\theta)= \mathbb{E}_{\eps\sim \mathcal{N}(\mathbf{0},\mathrm{I})}[f(\theta+\sigma \eps)] \ ,
\end{equation*}
where $\sigma$ is a parameter modulating the size of the smoothing area and $\mathcal{N}(\mathbf{0},\mathrm{I})$ is the $n$-dimensional Gaussian distribution with $\mathbf{0}$ being the all $0$ vector and $\mathrm{I}$ being the $n$-dimensional identity matrix.
The gradient of $f_{\sigma}$ with respect to parameters $\theta$ is given by 
\begin{equation*}
    \tri f_{\sigma} =\frac{1}{\sigma} \mathbb{E}_{\eps\sim \mathcal{N}(\mathbf{0},\mathrm{I})}[f(\theta+\sigma \eps)\eps] ,
\end{equation*}
which can be sampled by a Monte Carlo estimator, see~\cite{choromanski2018structured}. Often antithetic sampling is used, as it reduces variance~\cite{choromanski2018structured}. The \emph{antithetic ES gradient estimator} using $P$ samples is given by
\begin{equation}\label{eq:g-es}
    g_{ES}=  \frac{1}{2\sigma P}\sum _{i=1}^P \left(f(\theta+\sigma \eps^i)-f(\theta-\sigma \eps^i)\right)\eps ^i \ ,
\end{equation}{}
where $\eps ^i$ are independently sampled from $\mathcal{N}(\mathbf{0},\mathrm{I})$ for $i \in \{1,\ldots, P\}$. This gradient estimator has been shown to be effective in RL settings~\cite{salimans2017evolution}.




\subsection{Our One Step Gradient Estimator}
We first give some intuition before presenting our gradient estimator formally.
Given one surrogate gradient direction $\zeta$,  our one step gradient estimator applies the following sampling strategy. First, it estimates how much the gradient points into  the direction of $\zeta$ by antithetically evaluating $f$ in the direction of $\zeta$. Second, it estimates the part of $\tri f$ that is orthogonal to $\zeta$ by evaluating random, pairwise orthogonal search directions that are orthogonal to $\zeta$. In this way, our estimator detects the optimal lengths of the parameter update step into the both surrogate direction and the evaluated orthogonal directions (e.g.\ if $\zeta$ and $\tri f$ are parallel, the update step is parallel to $\zeta$, and if they are orthogonal the step into direction $\zeta$ has  length $0$). Additionally, if the  surrogate direction and the gradient are not perfectly aligned, then the gradient estimate almost surely improves over the surrogate direction due to the contribution from the evaluated directions orthogonal to $\zeta$. In the following we define our estimator formally and prove that the estimated direction possesses best possible alignment with the gradient that can be achieved with our sampling scheme.

We assume that $k$ pairwise orthogonal surrogate gradient directions $\zeta^1, \ldots, \zeta^k$ are given to our estimator. 
Denote by $\mathbb{R}_\zeta$ the subspace of $\mathbb{R}^n$ that is spanned by  the $\zeta^i$, and by $\mathbb{R}_{\perp \zeta}$ the subspace that is orthogonal to $\mathbb{R}_{\zeta}$. 
Further, for vectors $v$ and $\tri f$, we denote by $\hat{v}$ and $\htri f$ the normalized vector  $\frac{v}{\|v\|}$ and $\frac{\tri f}{\|\tri f\|}$, respectively.
Let $\heps^1, \ldots, \heps ^{P}$ be random orthogonal unit vectors from  $\mathbb{R}_{\perp \zeta}$. Then, our estimator is defined as
\begin{equation}\label{eq:our-estimator}
        g_{our}=  \sum _{i=1}^k \frac{f(\theta+\sigma \hzeta^i)-f(\theta-\sigma\hzeta^i)}{2\sigma}\hzeta^i+\sum _{i=1}^P  \frac{f(\theta+\sigma \heps^i)-f(\theta-\sigma\heps^i)}{2\sigma}\heps^i
         \ .
\end{equation}{}

We write $\tri f= \tri f_{\|\zeta}+ \tri f_{\perp \zeta}$, where  $\tri f_{\|\zeta}$ and  $\tri f_{\perp \zeta}$ are the projections of $\tri f$ on $\mathbb{R}_{\zeta}$ and $\mathbb{R}_{\perp\zeta}$, respectively.
In essence, the first sum in \eqref{eq:our-estimator} computes $\tri f _{\|\zeta}$ by assessing the quality of each surrogate gradient direction, and the second sum estimates $\tri f_{\perp \zeta}$ similar to an orthogonalized antithetic ES gradient estimator, that samples directions from $\mathbb{R}_{\perp \zeta}$, see~\cite{choromanski2018structured}.
We remark that we require pairwise orthogonal unit directions $\heps^i$ for the optimality proof. Due to the orthogonality of the directions, no normalization factor like the $   1/P$ factor in~\eqref{eq:g-es} is required in~\eqref{eq:our-estimator}.  In practice, the dimensionality $n$ is often much larger than $P$. Then, sampling pairwise orthogonal unit vecotrs $\eps^i$ is nearly identical to sampling the $\eps^i$s from a $\mathcal{N}(\mathbf{0},\mathrm{I})$ distribution, because in high-dimensional space the norm of $\eps^i\sim \mathcal{N}(\mathbf{0},\mathrm{I})$ is highly concentrated around $1$ and the cosine of two such random vectors is highly concentrated around $0$.

For the sake of analysis, we assume that $f$ is differentiable and we assume equality for the following first order approximation
$$
\frac{f(\theta+\sigma\heps) - f(\theta-\sigma\heps)}{2\sigma}\approx \ls \tri f(\theta), \heps \rs
$$

In the following, we will omit the $\theta$ in $\tri f(\theta)$.
Our first proposition states that $g_{our}$ computes the direction in the subspace spanned by $\zeta^1, \ldots, \zeta^k, \eps^1, \ldots, \eps^P$ that is most aligned with $\tri f$.

\begin{proposition}[Optimality of  $g_{our}$]\label{thm:optimal}
Let $\zeta^1, \ldots, \zeta^k, \eps^1, \ldots, \eps^P$ be pairwise orthogonal vectors in $\mathbb{R}^n$. Then, $g_{our}= \sum_{i=1}^k\ls\tri f , \hat{\zeta^i}\rs\hat{\zeta^i}+
\sum_{i=1}^P\ls\tri f, \hat{\eps^i}\rs\hat{\eps^i}$ computes the projection of $\tri f$ on the subspace spanned by $\zeta^1, \ldots, \zeta^k, \eps^1, \ldots, \eps^P$. Especially, $\eps = g_{our}$ is the vector of that subspace that maximizes the  cosine  $\ls\htri f ,\heps\rs$ between $\tri f$ and $\eps$. Moreover, the squared cosine between $g_{our}$ and $\tri f$ is given by \begin{equation}\label{eq:cosine-squared-many-directions}
    \ls\hat{\tri} f , \hat{g}_{our}\rs^2 = \sum_{i=1}^k\ls\hat{\tri} f , \hat{\zeta^i}\rs^2+\sum_{i=1}^P\ls\hat{\tri} f , \heps^i\rs^2 \ .
\end{equation}{}
\end{proposition}

We remark that when evaluating $\ls \tri f , v^i\rs$ for arbitrary directions $v^i$,  no information about search directions orthogonal to the subspace spanned by the $v^i$s is obtained. Therefore, one can only hope for finding the best approximation of $\tri f$ lying within the subspace spanned by the $v^i$s, which is accomplished by $g_{our}$. 
The proof of Proposition~\ref{thm:optimal}  follows easily from the Cauchy-Schwarz inequality and is given in the appendix.

\subsection{Iterative Gradient Estimation Using Past Descent Directions}
Our gradient estimation algorithm iteratively applies the one step gradient estimator $g_{our}$ by using the gradient estimate of the last time step as surrogate direction for the current time step. At any time step, our one step gradient estimator provides a better gradient estimate than the surrogate direction. Therefore, the algorithm accumulates information about the gradient, and the gradient estimate becomes more aligned with the true gradient over time.  Our algorithm relies on the assumption that gradients are correlated from one time step to the next.  This assumption is justified since it is one of the reasons momentum-based optimizers ~\cite{dozat2016incorporating, sutskever2013importance, ruder2016overview} are successful in deep learning. We also explicitly test this assumption experimentally, see Figure~\ref{fig:mnist_consecutive}. In the following, we first analyse how fast our algorithm accumulates information about the gradient over time if the gradient is constant, i.e.\ if a linear function is optimized. In this setting, we show that our algorithm is close to optimal, i.e.\ the convergence rate is only by a small constant factor smaller than the one of optimal orthogonal sampling, see Theorem~\ref{thm:convergence-rate-linear}.  Second, we analyse the quality of our gradient estimates for general, non-linear functions, where the gradient changes  over time. We show under some simplifying assumptions that, also in the general case, our algorithm builds up an improved gradient estimate over time, see Theorem~\ref{thm:convergence-rate-nonlinear}.


We first need some notation.
Denote by $\theta_t$ the search point, by $\tri f_t= \tri f(\theta_t)$ the gradient and by $\zeta_t$ the parameter update step at time $t$, that is, $\theta_{t+1}=\theta_t+\zeta_t$. 
The iterative gradient estimation algorithm obtains the gradient estimate $\zeta_t$ by  computing $g_{our}$ with the
  last update direction $\zeta_{t-1}$ as surrogate gradient and $P$ new random directions $  \heps^i$. Formally, let $
\heps^1_t, \ldots, \heps^P_t$ be pairwise orthogonal unit directions chosen uniformly from the unit sphere, that is, they are conditioned to be pairwise orthogonal and are marginally uniformly distributed. By defining $\eps_t= \sum_{i=1}^P \ls \tri f_t, \heps^i_t\rs \heps^i_t$, and setting $\zeta_t=g_{our}$, we obtain 

\begin{equation}\label{eq:iterative-estimation-scheme}
\zeta_{t}=  \ls \tri f_t, \hzeta_{t-1}\rs \hzeta_{t-1}+ \sum_{i=1}^P \ls \tri f_t, \heps_t^i\rs \heps_t^i = \ls \tri f_t, \hzeta_{t-1}\rs \hzeta_{t-1}+ \ls \tri f_t, \heps_t\rs \heps_t \ . 
\end{equation}{}
Then, Equation~\ref{eq:cosine-squared-many-directions} of Proposition~\ref{thm:optimal} turns into
\begin{equation}\label{eq:cosine-squared-one-direction}
    \ls \htri f_t, \zeta_{t} \rs^2= \ls \htri f_t, \hzeta_{t-1}\rs ^2+ \ls \htri f_t, \heps_t\rs^2 \ .
\end{equation}{}
The next theorem quantifies how fast the cosine between $\zeta_t$ and $\tri f_t$ converges to $1$, if  $\tri f_t$ does not change over time.

\begin{theorem}[Convergence rate for linear functions]\label{thm:convergence-rate-linear} Let $\zeta_t$ be iteratively computed using the past update direction  and $P$ pairwise orthogonal random directions, see Equation~\eqref{eq:iterative-estimation-scheme}, and
 let $X_t= \ls \htri f ,\hzeta_t\rs $ be the random variable that denotes the cosine between $\zeta_t$ and $\tri f_t$ at time $t$.
Then, the expected drift of $X_t^2$ is $\mathbb{E}[X_t^2- X_{t-1}^2|X_{t-1}=x_{t-1}]=  (1-x_{t-1}^2)\frac{P}{N-1} $. Moreover, let $\eps>0$ and  define  $T$ to be the first point in time $t$ with $X_t^2\geq 1 - \delta$.  It holds
\begin{equation*}
    \mathbb{E}[T]\leq  \frac{N-1}{P}\min \{ (1-\delta)/\delta,1+ \ln(1/ \delta)\} \ .
\end{equation*}{}
\end{theorem}{}

The first bound $\mathbb{E}[T]\leq\frac{N-1}{P} \frac{1-\delta}{\delta }$ is tight for $\delta$ close to $1$ and follows by an additive drift theorem, while the second bound $\mathbb{E}[T]\leq \frac{N-1}{P}(1+\ln (1/\delta))$ is tight for $\delta$ close to $0$ and follows by a variable drift theorem, see appendix. We remark that an optimal orthogonal gradient estimator requires $(1-\delta)N$ samples in order to reach a cosine squared of $1-\delta$. Since our algorithm evaluates $P+1$ directions per time step, it  requires approximately $\min\{1/\delta,\frac{1+\ln (1/\delta)}{1-\delta}\}$ times more samples to reach the same alignment.

Naturally, the linear case is not the most interesting one. However, it is hard to rigorously analyse the case of general $f$, because it is unpredictable how the gradient $\tri f_t$ differs from $\tri f_{t-1}$. Note that  $\tri f_t- \tri f_{t-1} \approx H \zeta_{t-1}$, where $H$ is the Hessian matrix of $f$ at $\theta_{t-1}$.
We define $\alpha_t = \ls \htri f_t, \htri f_{t-1}\rs$ and write $\htri f_t = \alpha_t\htri f_{t-1} + \tri f_{\perp}$ where $\tri f_{\perp}$ is orthogonal to $\htri f_t$ and has squared norm $1-\alpha_t^2$.
Then, the first term of \eqref{eq:cosine-squared-one-direction} is equal to
\begin{align*}
    \ls \htri f_t, \hzeta _{t-1}\rs ^2 
    = \ls \alpha_t\htri f_{t-1} + \tri f_{\perp}, \hzeta_{t-1}\rs^2
    = \left(\alpha_t \ls\htri f_{t-1}, \hzeta_{t-1}\rs + \ls\tri f_{\perp}, \hzeta_{t-1}\rs\right)^2 \ .
\end{align*}

In the following, we assume that $\tri f_{\perp}$ is a direction orthogonal to $\tri f_{t-1}$ chosen uniformly at random. Though, this assumption is not entirely true, it allows to get a grasp on the approximate cosine that our estimator is going to converge to.

\begin{theorem}\label{thm:convergence-rate-nonlinear}
Let $\zeta_t$ be iteratively computed using the past update direction  and $P$ pairwise orthogonal random directions, see Equation~\eqref{eq:iterative-estimation-scheme}, and
 let $X_t= \ls \htri f ,\hzeta_t\rs $ be the random variable that denotes the cosine between $\zeta_t$ and $\tri f_t$ at time $t$.
 Further, let $1\geq \alpha_ t \geq 0$ and assume that $\htri f_t = \alpha_{t}\htri f_{t-1}+ \tri f _{\perp}$, where $\tri f_{\perp}$ is a random vector orthogonal to $\htri f_{t-1}$ with norm $\sqrt{1-\alpha_t^2}$. 
 Choose $\zeta_t$ according to Equation~\eqref{eq:iterative-estimation-scheme} and define $X_t$ to be the cosine between $\htri f_t$ and $\hzeta_{t}$. Then,
\begin{equation*}
    \mathbb{E}[X_t^2|X_{t-1}=x_{t-1}] = \left(\alpha_t^2x_{t-1}^2 + (1-\alpha_t^2)(1-x_{t-1}^2)\frac{1}{N-1}\right)\left(1-\frac{P}{N-1}\right) + \frac{P}{N-1}\ .
\end{equation*}
\end{theorem}

The last theorem implies that the evolution of the cosine  depends heavily on the cosine $\alpha_t$ between consecutive gradients. Let $A= \frac{(1-\alpha_t^2) \frac{1}{N-1}(1-\frac{P}{N-1})+\frac{P}{N-1}}{1-(\alpha_t^2+ (1-\alpha_t^2\frac{1}{N-1})(1-\frac{P}{N-1}))}$. Then, the theorem implies that the drift $\mathbb{E}[X_t^2-X_{t-1}^2|X_{t-1}=x_{t-1}]$ is positive if $x_{t-1}\leq A$
and negative otherwise. Thus, if $\alpha_t$ would not change over time, we would expect $X_t$ to converge to $A$.


\section{Experiments}

In this section, we will empirically evaluate the performance of our gradient estimation scheme when combined with deep neural networks. In Section~\ref{sec:exp-mnist}, we show that it significantly improves gradient estimation for digit classifiers on MNIST. In Section~\ref{sec:exp-function-evaluation-noise}, we suggest how to overcome issues that arise from function evaluation noise. Finally, in Section~\ref{sec:exp-RL}, we evaluate our gradient estimation scheme on RL environments and investigate further issues arising in this setting.


\subsection{Gradient Estimation and Performance on MNIST} \label{sec:exp-mnist}

\begin{figure}
	\begin{subfigure}[b]{0.45\textwidth}
	\centering
		\includegraphics[width=1.0\linewidth]{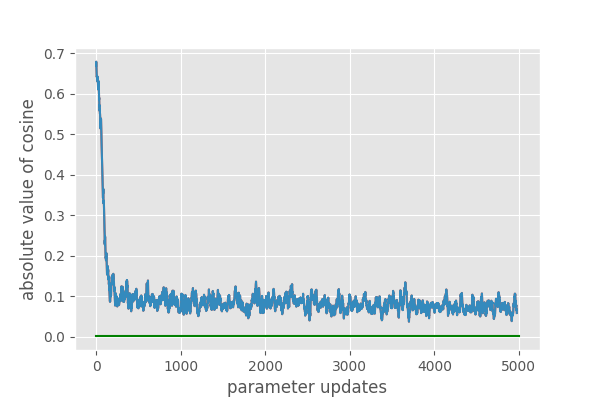}
		\caption{}
		\label{fig:mnist_consecutive}
    \end{subfigure}
    \hfill
    \begin{subfigure}[b]{0.45\textwidth}
		\centering
		\includegraphics[width=1.0\linewidth]{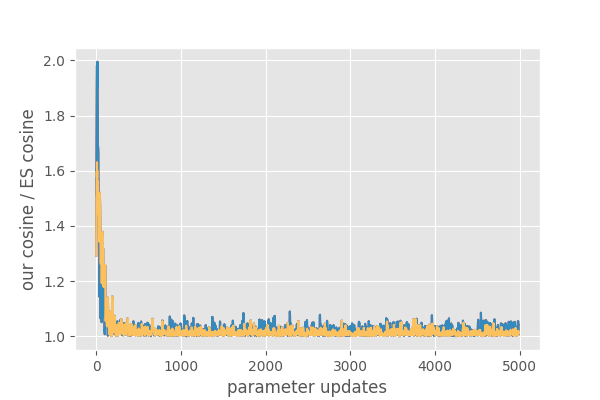}
		\caption{}
		\label{fig:mnist_ratio}
	\end{subfigure}%
	\caption{Improved gradient estimation. (a) The gradients before and after a parameter update are highly correlated. The  cosine between two consecutive gradients (blue line) and the cosine between two random vectors (green line) are plotted. (b) A network is trained with parameter updates according to $g_{ES}$ using SGD   (blue line) and Adam (yellow line) as optimizers. At any step we  compute our gradient estimate with $g_{our}$ and the true gradient $\bigtriangledown f$ with backpropagation. The plot shows that the ratio of the cosine between $g_{our}$ and $\bigtriangledown f $ and the cosine between $g_{ES}$ and $\bigtriangledown f$ is always strictly larger than $1$.}
	\label{fig:mnist-gradient-estimation}
\end{figure}

\begin{figure}[ht]
\vspace{-0.4cm}
        \centering
        
        \includegraphics[width=.85\linewidth]{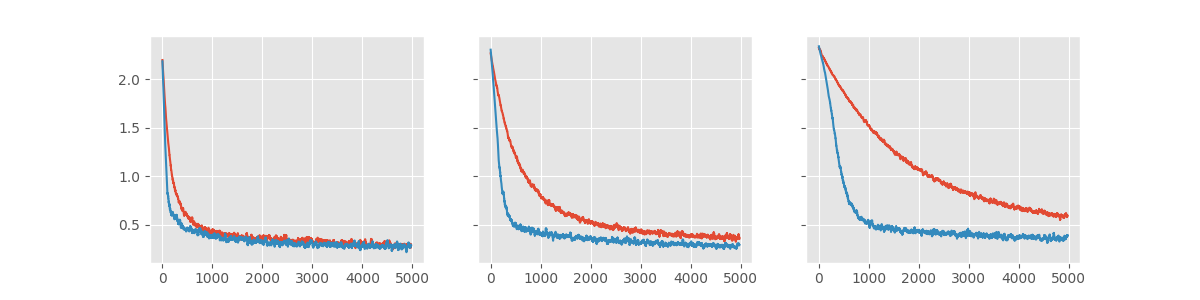}
        \includegraphics[width=.85\linewidth]{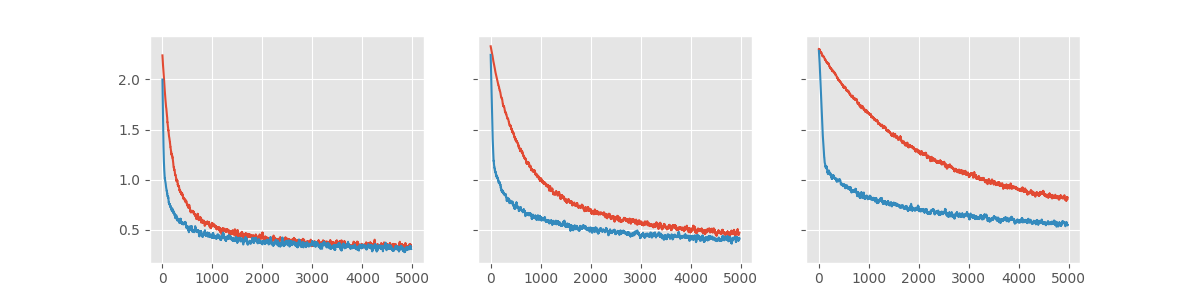}
        \caption[]{Performance of ES (red lines) and our algorithm (blue lines) on MNIST classification. The evolution of the training log-likelihood is plotted for the best three learning rates found for ES when using the Adam optimizer (top) and SGD (bottom). Our algorithm uses the same learning rates and hyper-parameters as ES.}
        \label{fig:mnist_results}
\vspace{-0.1cm}
\end{figure}

We observe that our approach significantly improves gradient estimation compared to standard ES. Figure~\ref{fig:mnist_consecutive} shows that the key requisite of our iterative gradient estimation scheme is satisfied during training on MNIST, that is, that gradients between consecutive parameter update steps are correlated.  Figure~\ref{fig:mnist_ratio} shows that our approach improves gradient estimation compared to ES during the whole training process and strongly improves it in the beginning of training, where consecutive gradients are most correlated, see Figure~\ref{fig:mnist_consecutive}.
 We observe that our approach  strongly outperforms ES in convergence speed and  reaches better final performance for all  hyperparameters we tested, see Figure~\ref{fig:mnist_results} and Table~\ref{table:mnist_results}. 

 \textbf{Implementation details:}
For these experiments, we  used a fully connected neural network with two hidden layers with a $tanh$ non-linearity and $1000$ units each, to have a high dimensional model ($\sim 1.8$ million parameters) .
For standard ES $128$ random search directions are evaluated at each step. For our algorithm the previous gradient estimate and $126$ random search directions are evaluated. We evaluated all directions on the same batch of images in order eliminate function evaluation noise and we  resampled after every update step.
We used small parameter perturbations ($\sigma = 0.001$). This is possible because no function evaluation noise is present and because the objective function is already differentiable and therefore no smoothing is required.
We test both SGD and Adam optimizers with learning rates in the range $10^{0.5}, 10^0, \ldots, 10^{-3}$.

\begin{table}[ht]
    \caption{
    Results on the MNIST digit classification task. We report the best loss and the number of steps until the training log-likelihood drops below 0.6, to observe the performance in the initial stages of learning. The values are for the best performing learning rate for each optimizer.}
    \label{table:mnist_results}
    \footnotesize
    \begin{center}
    \begin{tabular}{@{} lcc @{}} 
    \toprule[1.5pt]
        Optimizer   & Steps until loss $< 0.6$  & Best loss \\
        \midrule
        ES + Adam & 433 & 0.242 \\
        Ours + Adam & 182 & 0.216 \\
        \midrule
        ES + SGD & 727 & 0.305 \\
        Ours + SGD & 295 & 0.278 \\
    \bottomrule[1.5pt]
    \end{tabular}
    \end{center}
\end{table}


\subsection{Robustness to Function Evaluation Noise}\label{sec:exp-function-evaluation-noise}
\begin{figure}
	\begin{subfigure}[b]{0.3\textwidth}
	\centering
		\includegraphics[width=1.0\linewidth]{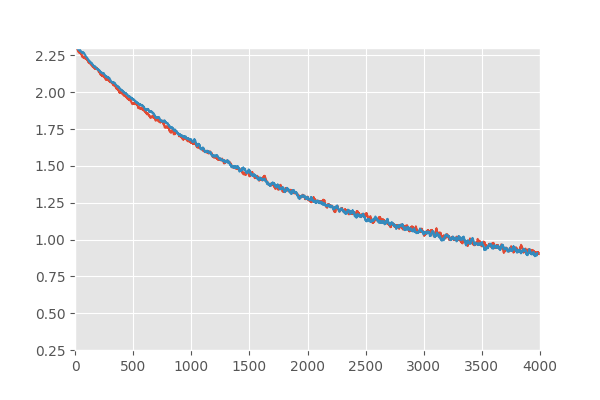}
		\caption{Default ES}
    \end{subfigure}
    \hfill
    \begin{subfigure}[b]{0.3\textwidth}
		\centering
		\includegraphics[width=1.0\linewidth]{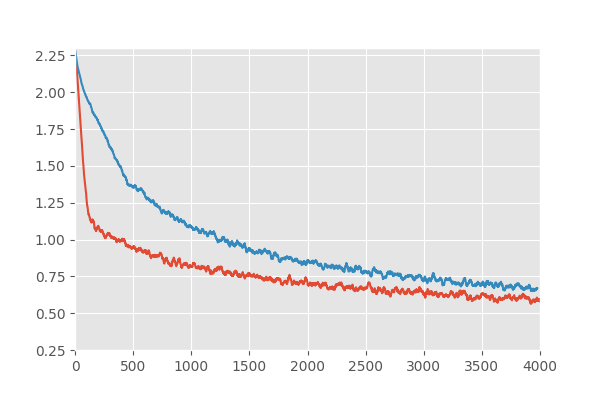}
		\caption{Ours with one past sample}
		\label{fig:mnist-noisy-1}
	\end{subfigure}%
	\hfill
    \begin{subfigure}[b]{0.3\textwidth}
		\centering
		\includegraphics[width=1.0\linewidth]{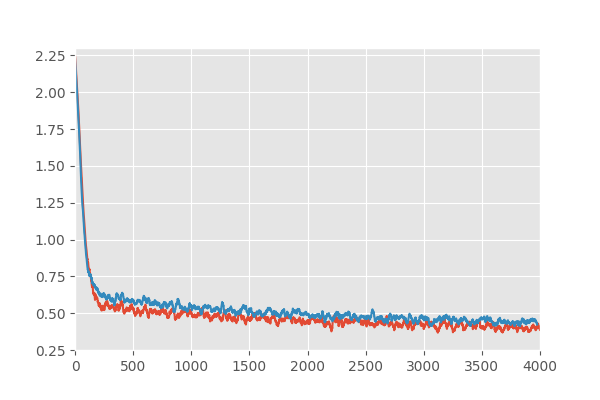}
		\caption{Ours with four past samples}
		\label{fig:mnist-noisy-4}
	\end{subfigure}%
	\caption{Using several past descent directions improves robustness to function evaluation noise. The plots show the performance on MNIST digit classification task with (blue lines) and without (red lines) function evaluation noise. The noise is created artificially by randomly permuting the fitness values of the evaluated search directions (see Equation~\ref{eq:our-estimator}) 
	 in $20\%$ of parameter updates. The x-axis represents the number of proper (i.e.\ non-permuted) parameter updates. (a) Standard ES does not suffer from this. (b) Function evaluation noise heavily impairs learning for our iterative gradient estimation scheme  when using one past update direction as surrogate gradient.
	 (c) Using $4$ past update directions as surrogate gradients makes our iterative gradient estimation scheme robust to function evaluation noise. }
	\label{fig:mnist-noisy}
\end{figure}

In practice, our iterative gradient estimation scheme may suffer from function evaluation noise because it builds up  good gradient estimates over several parameter update steps. Suppose that the past update direction is a good descent direction but it performs poorly on the current batch used for evaluation due to randomness in the batch selection or network evaluation process. Then, this direction is weighted lightly  when computing the new update direction, see Equation \ref{eq:iterative-estimation-scheme}, and therefore the information about this direction will be discarded. We empirically show, that our approach suffers heavily from this issue when artificially injecting noise in the function evaluation process, see Figure \ref{fig:mnist-noisy-1} . 
Figure \ref{fig:mnist-noisy-4} shows, that this issue can be resolved by using the last $k$ update directions for our gradient estimator (see Equation \ref{eq:our-estimator}). In this case, a good direction is only discarded, when it  performs poorly  in $k$ consecutive evaluation steps, which is very unlikely. We remark that the magnitude of the parameter updates naturally limits $k$, because the $k$-th last update direction is only useful if it is still  correlated with the current gradient. Concretely, we found that using the last $4$ parameters updates was extremely helpful for smaller learning rates, even in the absence of noise (see Figure \ref{fig:mnist-noisy-4}). However, it did not offer an advantage for larger learning rates. 

\subsection{Robotic RL environments}\label{sec:exp-RL}

\begin{figure}[ht]
        \centering
        \includegraphics[width=.85\linewidth]{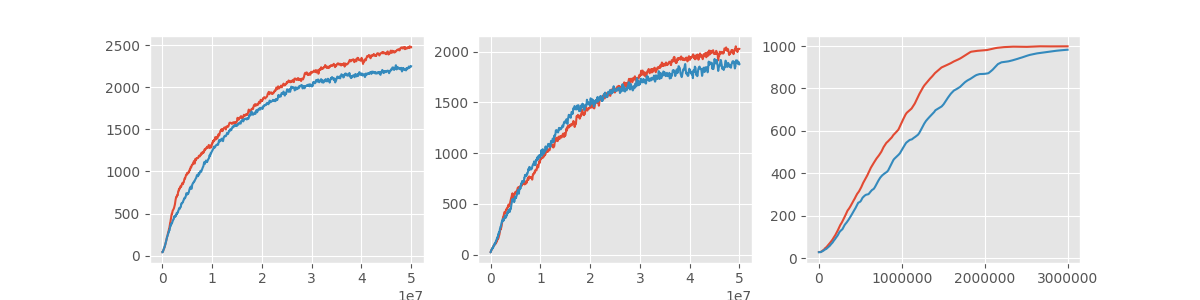}
        \caption[]{Performance of our algorithm (red line) and ES (blue line) on three different Roboschool tasks:  Ant (left), Cheetah (center) and Pendulum (right). The plot shows the mean average reward over $9$ repetitions as a function of time-steps (in thousands).}
        \label{fig:RL_reward}
\vspace{-0.1cm}
\end{figure}

For the next set of experiments, we evaluate our algorithm on three  robotics tasks of the Roboschool environment: RoboschoolInvertedPendulum-v1, RoboschoolHalfCheetah-v1 and RoboschoolAnt-v1. 
Our approach outperforms ES in the pendulum task, and offers a small improvement over ES in the other two tasks, see
Figure~\ref{fig:RL_reward}. The improvement of our approach over standard ES is smaller on RL tasks than on the MNIST task. Therefore, we first empirically confirmed that past updated direction are also in RL correlated with the gradient. To test this, we kept track of the average difference between random perturbations and the direction given by our algorithm, after normalizing the rewards. We found that, the direction of our algorithm had an average weight of $1.11$ versus the $0.65$ of a random direction.

RL robotics tasks bring two additional major challenges compared to the MNIST task. First, exploration is crucial to escape local optima and find new solutions, and second, the function evaluation noise is huge due to each perturbation being tested only on a single trajectory. Our proposed solution of robustness  against function evaluation noise intertwines with the exploration issue. A rather small step size is necessary in order to use more past directions as surrogate gradients. However, exploration in ES is driven by large perturbation sizes and noisy optimization trajectories. We did not observe improvements when combining the approach of using several past directions with standard hyperparameter settings. We believe that an exhaustive empirical study can shed light onto the effect of our approach on exploration and may further improve the performance on RL tasks. However, running extensive experiments for complex RL environment is computationally expensive.

\textbf{Implementation details:}  We use most of the hyper-parameters from  the OpenAI implementation
\footnote{\url{https://github.com/openai/evolution-strategies-starter}}
. That is, two hidden layers of $256$ units each and with a $tanh$ non-linearity. Further, we use a learning rate of $0.01$, a perturbation standard deviation of $\sigma = 0.02$ and the Adam optimizer, and we also apply fitness shaping \cite{wierstra2014natural}. 
For standard ES $128$ random perturbations are evaluated at each step. For our algorithm the previous gradient estimate and $126$ random perturbations are evaluated.
For the Ant and Cheetah environments, we observed with this setup, that  agents often get stuck in a local optima where they stay completely still, instead of running forward. As this happens for both,  ES and our algorithm, we tweaked the environments in order to ensure that a true  solution to the task is learned and not some  some degenerate optima, we tweaked the environments in the following way. We remove the penalty for using electricity and finish the episode if the agent does not make any progress in a given amount of time. In this way, agents  consistently escape the local minima. We use a $tanh$ non-linearity on the output of the network, which increased stability of training, as otherwise the output of the network would become very large without an electricity penalty.

\section{Conclusion}

We proposed a gradient estimator that optimally incorporates surrogate gradient directions and random search directions, in the sense that it determines the direction  with maximal cosine to the true gradient from the subspace of evaluated directions. Such a method has many applications as elucidated in~\cite{maheswaranathan2019guided}. Importantly, our estimator does not require information about the quality of surrogate directions, which allows  us to iteratively  use past update directions as surrogate directions for our gradient estimator. We theoretically quantified the benefits of the proposed iterative gradient estimation scheme. Finally, we showed that our approach in combination with deep neural networks considerably improves the gradient estimation capabilities of ES, at no extra computational cost.
The results on MNIST indicate that the speed of the Evolutionary Strategies themselves, a key part in the current Reinforcement Learning toolbox, is greatly improved.
Within Reinforcement Learning an out of the box application of our algorithm yields some improvements. The smaller improvement in RL compared to MNIST is likely due to the interaction of our approach and exploration that is essential in RL. We leave it to future work to explicitly add and study appropriate exploration strategies which might unlock the true potential of our approach in RL.

\bibliographystyle{unsrt}  
\bibliography{references}  

\begin{thebibliography}{10}

\bibitem{rechenberg1973evolution}
Ingo Rechenberg.
\newblock Evolution strategy: Optimization of technical systems by means of
  biological evolution.
\newblock {\em Fromman-Holzboog, Stuttgart}, 104:15--16, 1973.

\bibitem{schwefel1977evolutionsstrategien}
Hans-Paul Schwefel.
\newblock Evolutionsstrategien f{\"u}r die numerische optimierung.
\newblock In {\em Numerische Optimierung von Computer-Modellen mittels der
  Evolutionsstrategie}, pages 123--176. Springer, 1977.

\bibitem{nesterov2017random}
Yurii Nesterov and Vladimir Spokoiny.
\newblock Random gradient-free minimization of convex functions.
\newblock {\em Foundations of Computational Mathematics}, 17(2):527--566, 2017.

\bibitem{salimans2017evolution}
Tim Salimans, Jonathan Ho, Xi~Chen, Szymon Sidor, and Ilya Sutskever.
\newblock Evolution strategies as a scalable alternative to reinforcement
  learning.
\newblock {\em arXiv preprint arXiv:1703.03864}, 2017.

\bibitem{choromanski2018structured}
Krzysztof Choromanski, Mark Rowland, Vikas Sindhwani, Richard~E Turner, and
  Adrian Weller.
\newblock Structured evolution with compact architectures for scalable policy
  optimization.
\newblock {\em arXiv preprint arXiv:1804.02395}, 2018.

\bibitem{cui2018evolutionary}
Xiaodong Cui, Wei Zhang, Zolt{\'a}n T{\"u}ske, and Michael Picheny.
\newblock Evolutionary stochastic gradient descent for optimization of deep
  neural networks.
\newblock In {\em Advances in neural information processing systems}, pages
  6048--6058, 2018.

\bibitem{houthooft2018evolved}
Rein Houthooft, Yuhua Chen, Phillip Isola, Bradly Stadie, Filip Wolski,
  OpenAI~Jonathan Ho, and Pieter Abbeel.
\newblock Evolved policy gradients.
\newblock In {\em Advances in Neural Information Processing Systems}, pages
  5400--5409, 2018.

\bibitem{ha2018recurrent}
David Ha and J{\"u}rgen Schmidhuber.
\newblock Recurrent world models facilitate policy evolution.
\newblock In {\em Advances in Neural Information Processing Systems}, pages
  2450--2462, 2018.

\bibitem{metz2018learned}
Luke Metz, Niru Maheswaranathan, Jeremy Nixon, Daniel Freeman, and Jascha
  Sohl-dickstein.
\newblock Learned optimizers that outperform on wall-clock and validation loss.
\newblock 2018.

\bibitem{bengio2013estimating}
Yoshua Bengio, Nicholas L{\'e}onard, and Aaron Courville.
\newblock Estimating or propagating gradients through stochastic neurons for
  conditional computation.
\newblock {\em arXiv preprint arXiv:1308.3432}, 2013.

\bibitem{watkins1992q}
Christopher~JCH Watkins and Peter Dayan.
\newblock Q-learning.
\newblock {\em Machine learning}, 8(3-4):279--292, 1992.

\bibitem{rumelhart1985learning}
David~E Rumelhart, Geoffrey~E Hinton, and Ronald~J Williams.
\newblock Learning internal representations by error propagation.
\newblock Technical report, California Univ San Diego La Jolla Inst for
  Cognitive Science, 1985.

\bibitem{lillicrap2014random}
Timothy~P Lillicrap, Daniel Cownden, Douglas~B Tweed, and Colin~J Akerman.
\newblock Random feedback weights support learning in deep neural networks.
\newblock {\em arXiv preprint arXiv:1411.0247}, 2014.

\bibitem{maheswaranathan2019guided}
Niru Maheswaranathan, Luke Metz, George Tucker, Dami Choi, and Jascha
  Sohl-Dickstein.
\newblock Guided evolutionary strategies: augmenting random search with
  surrogate gradients.
\newblock In {\em International Conference on Machine Learning}, pages
  4264--4273, 2019.

\bibitem{dozat2016incorporating}
Timothy Dozat.
\newblock Incorporating nesterov momentum into adam.
\newblock 2016.

\bibitem{sutskever2013importance}
Ilya Sutskever, James Martens, George Dahl, and Geoffrey Hinton.
\newblock On the importance of initialization and momentum in deep learning.
\newblock In {\em International conference on machine learning}, pages
  1139--1147, 2013.

\bibitem{ruder2016overview}
Sebastian Ruder.
\newblock An overview of gradient descent optimization algorithms.
\newblock {\em arXiv preprint arXiv:1609.04747}, 2016.

\bibitem{hansen2016cma}
Nikolaus Hansen.
\newblock The cma evolution strategy: A tutorial.
\newblock {\em arXiv preprint arXiv:1604.00772}, 2016.

\bibitem{wierstra2014natural}
Daan Wierstra, Tom Schaul, Tobias Glasmachers, Yi~Sun, Jan Peters, and
  J{\"u}rgen Schmidhuber.
\newblock Natural evolution strategies.
\newblock {\em The Journal of Machine Learning Research}, 15(1):949--980, 2014.

\bibitem{kingma2014adam}
Diederik~P Kingma and Jimmy Ba.
\newblock Adam: A method for stochastic optimization.
\newblock {\em arXiv preprint arXiv:1412.6980}, 2014.

\bibitem{lengler2018drift}
Johannes Lengler and Angelika Steger.
\newblock Drift analysis and evolutionary algorithms revisited.
\newblock {\em Combinatorics, Probability and Computing}, 27(4):643--666, 2018.

\end{thebibliography}

\appendix
\section{Proof of Theorems}
In this Section we prove the theorems from the main paper rigorously.

\subsection{Proof of Proposition \ref{thm:optimal}}
Note that for this theorem there is no distinction between the directions $\zeta^i$ and $\eps^i$. For ease of notation, we denote $\zeta^1, \ldots, \zeta^k, \zeta^1, \ldots, \zeta^P$ by $\eps^1, \ldots, \eps^m$.  The theorem is a simple application of the Cauchy-Schwarz inequality.  Denote by $\tri f_{\|\eps^i}= \sum_{i=1}^m\ls\tri f, \hat{\eps}^i\rs\hat{\eps}^i$ the projection of $\tri f$ on the subspace spanned by the $\eps^i$s, and let   $\eps = \sum_{i=1}^m \alpha_i\hat{\eps}^i$ be a vector in that subspace.
Then, the Cauchy-Schwarz inequality implies 
\begin{equation}
\ls\tri f, \eps\rs = \ls\tri f_{\|\eps^i}, \eps\rs\leq \|\tri f_{\|\eps^i}\|\|\eps\| \ .
\end{equation}
 Equality holds if and only if $\eps$ and $\tri f_{\|\eps^i}$ have the same direction, which is equivalent to $\eps= \alpha g_{our}$ for some $\alpha>0$. In particular, in this case the cosine squared between   $\tri f$ and $\eps$ is 
\begin{equation}
\ls\hat{\tri} f, \hat{\eps}\rs^2 = \frac{\|\tri f_{\|\eps^i}\|^2}{\|\tri f\|^2} = \sum_{i=1}^m\ls\hat{\tri} f, \hat{\eps}^i\rs^2
\end{equation}\qed

\subsection{Expectation of Cosine Squared of Random Vectors from the Unit Sphere}
We need the following proposition for the proofs of Theorem~\ref{thm:convergence-rate-linear} and~\ref{thm:convergence-rate-nonlinear}. 

\begin{proposition}\label{prop:cosine-unit-vectors}
Let $\hat{u}$ be an $N$-dimensional unit vector, and let $\heps^1, \ldots, \heps^P$ be pairwise orthogonal vectors sampled uniformly from the $N$-dimensional unit sphere, that is, they are marginally uniformly distributed and conditioned to be pairwise orthogonal. Then, the expected cosine squared of $\hat{u}$ and $\eps= \sum_{i=1}^P \ls u, \heps^i \rs \heps^i$ is
\begin{equation*}
    \mathbb{E}[\ls \hat{u}, \heps \rs^2]= \frac{P}{N}
\end{equation*}{}
\end{proposition}{}

\begin{proof}
Note that 
\begin{equation*}
    \ls \hat{u}, \heps \rs^2 = \frac{1}{\|\eps\|^2}\ls \hat{u}, \eps \rs^2= \frac{1}{\sum_{i=1}^P \ls \hat{u}, \heps^i\rs ^2}\left(\sum_{i=1}^P \ls \hat{u}, \heps^i\rs ^2\right)^2 = \sum_{i=1}^P \ls \hat{u}, \heps^i\rs ^2
\end{equation*}{}

Denote by $S$ the $N$-dimensional unit sphere. Linearity of expectation implies that
\begin{align*}
    \mathbb{E}[\ls \hat{u}, \heps \rs^2]&= \sum_{i=1}^P \mathbb{E}[\ls \hat{u}, \heps^i\rs ^2]\\
    & = \frac{P}{Vol(S)} \int_S \ls \hat{u}, v\rs^2\  \mathrm{d}v
\end{align*}{}
By rotational invariance of the unit sphere, we can replace $\hat{u}$ by $e_1=(1, 0, \ldots, 0)$ and obtain
\begin{align*}
    \mathbb{E}[\ls \hat{u}, \heps \rs^2]&= \frac{P}{Vol(S)} \int_S \ls e_1, v\rs^2\ \mathrm{d}v \\
    &=\frac{P}{Vol(S)} \int_S v_1^2\  \mathrm{d}v \\
    &= \frac{P}{N \cdot Vol(S)} \int_S \sum_{i=1}^N v_i^2 \ \mathrm{d}v \\
        &= \frac{P}{N \cdot Vol(S)} \int_S 1 \ \mathrm{d}v \\
    &= \frac{P}{N}
\end{align*}{}

\end{proof}{}


\subsection{Proof of Theorem \ref{thm:convergence-rate-linear}}
In order to prove Theorem~\ref{thm:convergence-rate-nonlinear}, use Equation~\ref{eq:cosine-squared-one-direction} to compute how $X_{t}^2=\ls \htri f, \hzeta_t \rs^2$ depends on $X_{t-1}^2$. Then, we apply a variable transformation to $X_t$ in order to be able to apply the additive and variable drift theorems from~\cite{lengler2018drift}, which are stated in Section~\ref{sec:drift-theorems}.

We can split the normalized gradient $\htri f= \tri f _{\perp\hzeta_{t-1}} +\tri f _ { \|\hzeta_{t-1}}$ into an orthogonal to $\hzeta_{t-1}$ part and a parallel to $\hzeta_{t-1}$ part.  
It holds  $\| \tri f _ {\perp\hzeta_{t-1}} \|^2= 1-\ls \htri f, \hzeta_{t-1}\rs^2$ and $\ls \htri f_{\|\hzeta_{t-1}},\heps_t \rs=0$ since $\heps_t$ is a  unit vector orthogonal to $\hzeta_{t-1}$. Recall that $\htri f_{\perp \zeta}=\frac{\tri f_{\perp \zeta}}{\|\tri f_{\perp \zeta}\|}$, then
\begin{align}\label{eq:eps-direction}
   \ls \htri f, \heps_t\rs ^2 &= \ls \tri f_ {\perp\hzeta_{t-1}}, \heps_t\rs^2 = (1-\ls \htri f, \hzeta_{t-1}\rs^2 ) \ls \htri f_{\perp \zeta}, \heps_t \rs ^2 = (1-X_{t-1}^2) \ls \htri f_{\perp \zeta}, \heps_t \rs^2\ ,
\end{align}
and therefore by Equation~\ref{eq:cosine-squared-one-direction}
\begin{align}
        X_t^2&= \ls \htri f, \hzeta _{t-1}\rs ^2 + \ls \htri f, \heps _t\rs ^2  
        =  X_{t-1}^2+ (1-X_{t-1}^2 ) \ls \htri f_{\perp \zeta}, \heps_t \rs ^2 \ . 
\end{align}
Define the random process $Y_t=1-X_t^2$. 
It holds 
\begin{align}
    &Y_t= 1-X_{t-1}^2 + (1-X_{t-1}^2)\ls \htri f_{\perp \zeta}, \heps_t \rs ^2= Y_{t-1}(1-\ls \htri f_{\perp \zeta}, \heps_t \rs ^2)\ ,\
    \text{and}  \\
    & \mathbb{E}[Y_t| Y_{t-1}=y_{t-1}]= y_{t-1}\left(1-  \mathbb{E}[\ls \htri f_{\perp \zeta}, \heps_t \rs ^2] \right)= y_{t-1}\left(1-\frac{P}{N-1}\right) \ , \label{eq:drift-yt}
\end{align}{}
where we used Proposition~\ref{prop:cosine-unit-vectors} in the $N-1$ dimensional subspace that is orthogonal to $\zeta_{t-1}$.

In order to derive the first bound on $T$, we   bound the drift of $Y_t$ for $Y_t\geq \delta$.

\begin{equation*}
    \mathbb{E}[Y_t| Y_{t-1}=y_{t-1}, y_{t-1}\geq \delta]=  y_{t-1}- y_{t-1}\frac{P}{N-1} \leq y_{t-1}-\delta \frac{P}{N-1}\ , 
\end{equation*}{}
where we used Equation~\ref{eq:drift-yt} and $y_{t-1} \geq \delta$.
In order to apply Theorem~\ref{thm:additive-drift}, we define the auxiliary process $Z_t=Y_t-\delta$. Then, $T$ is the expected time that $Z_t$ hits $0$. Since $\mathbb{E}[Z_t| Z_{t-1}=z_{t-1}, z_{t-1}\geq 0]   \leq z_{t-1}-\delta \frac{c}{N-1}$ and  $Z_0=1-\delta$,  Theorem~\ref{thm:additive-drift} implies that 
\begin{equation*}
    \mathbb{E}[T]\leq \frac{1-\delta}{\delta}\frac{N-1}{P}
\end{equation*}{}


In order to apply Theorem~\ref{thm:general-drift}, to show the second bound on $T$, we need to rescale $Y_t$ such that it takes values in $\{0\}\cup[1,\infty)$. Define the auxiliary process $Z_t$ by 
\begin{equation}\label{eq:def-Zt}
    Z_t = \begin{cases}
    Y_t/\delta & \text{ if }  Y_t\geq \delta \\
    0 & \text{ if } Y_t<\delta 
    \end{cases}\ . 
\end{equation}{}
Then, $T$ is the expected time that $Z_t$ hits $0$.
The process $Z_t$ satisfies
\begin{equation*}
    \mathbb{E}\left[Z_t|Z_{t-1}=z_{t-1}, z_{t-1}\geq 1\right] \leq \mathbb{E}\left[Y_t/\delta| Y_{t-1}=\delta z_{t-1}, z_{t-1}\geq 1\right] \leq z_{t-1}\left(1-\frac{P}{N-1}\right) \ ,
\end{equation*}{}
where we used Equations \ref{eq:def-Zt} and \ref{eq:drift-yt}.
Since $Z_0=1/\delta$,  Theorem~\ref{thm:general-drift}  implies for  $h(z)=z\frac{c}{N-1}$  that
\begin{equation*}
    \mathbb{E}[T]\leq \frac{N-1}{P} + \int _1^{1/\delta} \frac{N-1}{P u } \ \mathrm{d}u = \frac{N-1}{P}(1+ \ln (1/\delta)) \ .
\end{equation*}{}
\qed

\subsection{Proof of Theorem \ref{thm:convergence-rate-nonlinear}}

In order to prove the theorem, we need to understand how $X_t$ depends on the value of $X_{t-1}$. It holds $X_t ^2=\ls \htri f_t, \hat{\zeta}_{t-1}\rs ^2 + \ls \htri f_t, \hat{\eps}_{t}\rs ^2$. As in Equation~\ref{eq:drift-yt}, we can write $\ls \htri f_t, \hat{\eps}_{t}\rs ^2 = (1-\ls \htri f_t, \hat{\zeta}_{t-1}\rs ^2)\ls \htri f_{\perp\hzeta_{t-1}}, \hat{\eps}_{t}\rs ^2$, and note that $\mathbb{E}[\ls \htri f_{\perp \hzeta_{t-1}}, \heps_t\rs^2]= \frac{P}{N-1}$ because $\heps$ is a random direction orthogonal to $\hzeta_{t-1}$. 
This implies that
\begin{align}\label{eq:1}
    \mathbb{E}[X_t^2|X_{t-1}=x_{t-1}] &= \left(1-\frac{P}{N-1}\right)\mathbb{E}[\ls \htri f_t, \hat{\zeta}_{t-1}\rs ^2|X_{t-1}=x_{t-1}] + \frac{P}{N-1}
\end{align}{}

 To understand how the $X_t$ evolves we need to analyze how $\ls \htri f_t, \hat{\zeta}_{t-1}\rs ^2$ relates to $X_{t-1}=\ls \htri f_{t-1}, \hat{\zeta}_{t-1}\rs ^2$.
To that end, we set $\alpha_t = \ls \htri f_t, \htri f_{t-1}\rs$ and write $\htri f_t = \alpha_t\htri f_{t-1} + \tri f_{\perp\tri f_{t-1}}$ where $\tri f_{\perp\tri f_{t-1}}$ is orthogonal to $\tri f_{t-1}$ and has norm $1-\alpha_t^2$.
Then, 
\begin{align*}
    \ls \htri f_t, \hzeta _{t-1}\rs ^2 
    = \ls \alpha_t\htri f_{t-1} + \tri f_{\perp\tri f_{t-1}}, \hzeta_{t-1}\rs^2
    = \left(\alpha_t \ls\htri f_{t-1}, \hzeta_{t-1}\rs + \ls\tri f_{\perp\tri f_{t-1}}, \hzeta_{t-1}\rs\right)^2 \ .
\end{align*}

It follows that 
\begin{align}
& \mathbb{E}[ \ls \htri f_t, \hzeta _{t-1}\rs ^2 |X_{t-1}=x_{t-1}]\\
&= \alpha_t^2 x_{t-1}^2 + 2 \alpha_tx_{t-1} \mathbb{E}[\ls \tri f_{\perp\tri f_{t-1}}, \hzeta_{t-1}\rs]+ \mathbb{E}[\ls \tri f_{\perp\tri f_{t-1}}, \hzeta_{t-1}\rs^2]\\
&= \alpha_t^2 x_{t-1}^2 +(1-\alpha^2_t)(1-x_t^2) \frac{1}{N-1} \ , \label{eq:2}
\end{align}{}
where we used $\mathbb{E}[\ls \tri f_{\perp\tri f_{t-1}},\hzeta_{t-1}\rs]=0$, which follows from the assumption of $\tri f_{\perp\tri f_{t-1}}$ being a random direction orthogonal to $\tri f_{t-1}$, and that
\begin{align*}
    \mathbb{E}[\ls \tri f_{\perp\tri f_{t-1}}, \hzeta_{t-1}\rs^2]
    &=\mathbb{E}[\ls \tri f_{\perp\tri f_{t-1}}, \hzeta_{t-1\perp \tri f_{t-1}}\rs^2]\\
    &=\|\tri f_{\perp\tri f_{t-1}}\|^2 \|\hzeta_{t-1\perp \tri f_{t-1}}\|^2 \mathbb{E}[\ls \frac{\tri f_{\perp\tri f_{t-1}}}{\| \tri f_{\perp\tri f_{t-1}}\|}, \frac{\hzeta_{t-1\perp \tri f_{t-1}}}{\|\hzeta_{t-1\perp \tri f_{t-1}}\|}\rs^2]\\
    &=(1-\alpha^2_t)(1-x_t^2)\frac{1}{N-1} \ ,
\end{align*}{}
which follows from $\| \tri f_{\perp\tri f_{t-1}}\|^2=1-\alpha^2$, $\|\hzeta_{t-1\perp \tri f_{t-1}}\|^2=1-x_t^2$ and Proposition~\ref{prop:cosine-unit-vectors} for $P=1$ using that $\tri f_{\perp\tri f_{t-1}}$ is a random direction orthogonal to $\tri f_{t-1}$.
Then, plugging  Equation~\eqref{eq:2} into \eqref{eq:1}, implies the theorem.\qed

\section{Drift Theorems}\label{sec:drift-theorems}
For the proof of Theorem~\ref{thm:convergence-rate-linear}, we use two drift theorems from~\cite{lengler2018drift}, which we restate for completeness.
\begin{theorem}[Additive Drift, Theorem $1$ from \cite{lengler2018drift}] \label{thm:additive-drift}
Let $(X_t)_{t\in \mathbb{N}_0}$ be a Markov chain with state space $S\subset [0, \infty)$ and assume $X_0=n$. Let $T$ be the earliest point in time $t\geq 0$ such that $X_t=0$. If there exists $c>0$  such that for all $x \in S$,  $x>0$ and for all $t\geq 0$ we have 
\begin{equation*}
    \mathbb{E}[X_{t+1}|X_t=x]\leq x-c \ .
\end{equation*}{}
Then,
\begin{equation*}
    \mathbb{E}[T] \leq \frac{n}{c} \ . 
\end{equation*}{}
\end{theorem}{}
\begin{theorem}[]Variable Drift, Theorem $4$ from \cite{lengler2018drift}]\label{thm:general-drift}
Let $(X_t)_{t\in \mathbb{N}}$ be a Markov chain with state space $S \subset \{0\}\cup [1,\infty)$ and with $X_0=n$. Let $T$ be the earliest point in time $t\geq 0$ such that $X_t= 0$. Suppose furthermore that there is a positive, increasing function $h:[1,\infty) \rightarrow \mathbb{R}_{>0}$ such that for all $x\in S$, $x>0$ we have for all $t\geq0$

\begin{equation*}
    \mathbb{E}[X_{t+1}|X_t=x]\leq x-h(x) \ .
\end{equation*}{}
Then, 
\begin{equation*}
    \mathbb{E}[T]\leq \frac{1}{h(1)}+ \int_1^n \frac{1}{h(u)}\ \mathrm{du} \ .
\end{equation*}{}
\end{theorem}{}{}

\end{document}